\documentclass[10pt,twocolumn,letterpaper]{article}

\usepackage{cvpr}              
\usepackage[accsupp]{axessibility}
\usepackage[dvipsnames]{xcolor}

\newcommand{\myparagraph}[1]{\vspace{1pt} \noindent \textbf{#1} \ }

\definecolor{cvprblue}{rgb}{0.21,0.49,0.74}
\usepackage[pagebackref,breaklinks,colorlinks,citecolor=cvprblue]{hyperref}
\usepackage{url}
\usepackage{graphicx}
\usepackage{subcaption}
\usepackage{multirow}
\usepackage{arydshln}

\usepackage{dsfont}

\def\paperID{8358} 
\def\confName{CVPR}
\def\confYear{2024}

\title{Attention Calibration for Disentangled Text-to-Image Personalization}

\author{Yanbing Zhang$^{1,2}$, Mengping Yang$^{1,2}$, Qin Zhou$^{1,2*}$, Zhe Wang$^{1,2*}$\\
\textsuperscript{1} Department of Computer Science and Engineering, ECUST, China\\
\textsuperscript{2} Key Laboratory of Smart Manufacturing in Energy Chemical Process, ECUST, China\\
{\tt\small \{zhangyanbing, mengpingyang\}@mail.ecust.edu.cn, \{sunniezq, wangzhe\}@ecust.edu.cn}
}

\begin{document}

\twocolumn[{%
\renewcommand\twocolumn[1][]{#1}%
\maketitle

\begin{center}
    \centering
    \includegraphics[width=\linewidth]{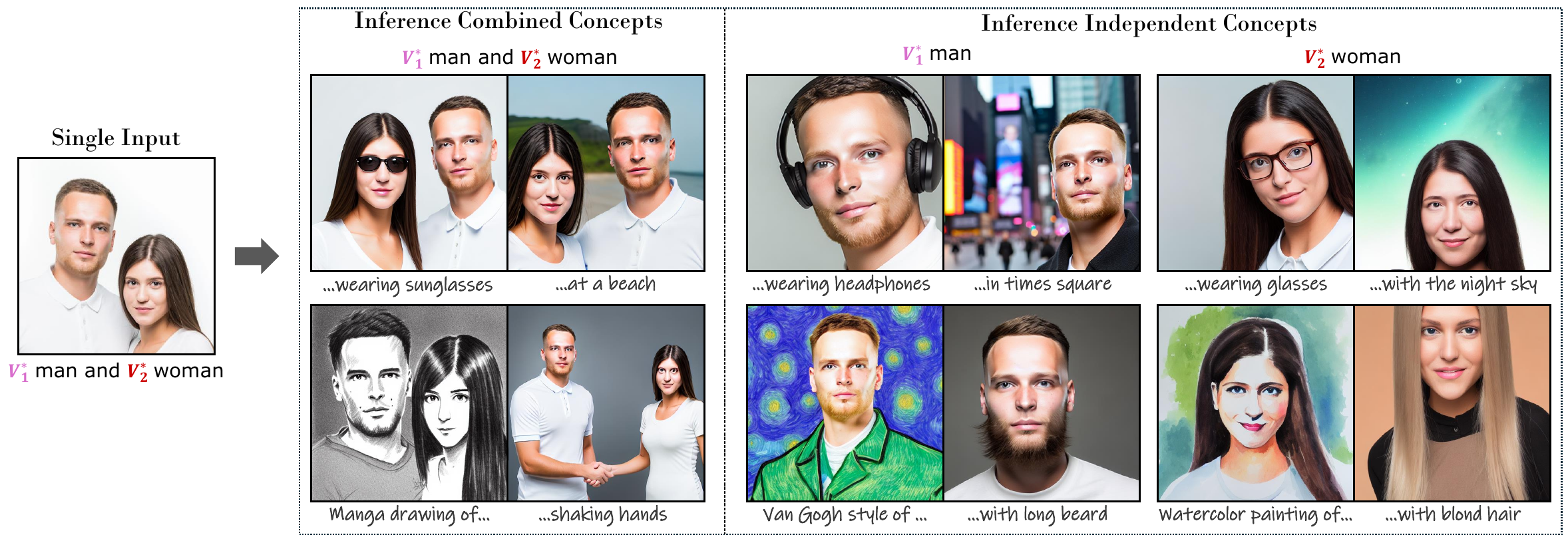}
    \vspace{-7pt}
\captionof{figure}{
Given one \emph{individual} image from specific users, our proposed method is capable of producing \emph{customized} images for each concept contained in the input image,
\emph{e.g.,} given a single input image with a man and a woman, our method excels in achieving innovative renditions of both combined (\emph{left}) and independent (\emph{right}) concepts, without compromising the fidelity and identity preservation, and more importantly, manifesting satisfactory interactive generation conditioned by various text prompts.
Note that we employ notation $V_i^*$ to denote the modifier of the $i$-th concept.
Our code and data will be publicly available at: \href{https://github.com/Monalissaa/DisenDiff}{https://github.com/Monalissaa/DisenDiff}.
}
    \label{fig:teaser}
\end{center}

}]

\maketitle

{
  \renewcommand{\thefootnote}%
    {\fnsymbol{footnote}}
  \footnotetext[1]{Corresponding author}
}

\begin{abstract}
    Recent thrilling progress in large-scale text-to-image (T2I) models has unlocked unprecedented synthesis quality of AI-generated content (AIGC) including image generation, 3D and video composition.
    Further, personalized techniques enable appealing customized production of a novel concept given only several images as reference.
    However, an intriguing problem persists: Is it possible to capture \textbf{multiple, novel concepts} from \textbf{one single reference image}? 
    In this paper, we identify that existing approaches fail to preserve visual consistency with the reference image and eliminate cross-influence from concepts.
    To alleviate this, we propose an attention calibration mechanism to improve the concept-level understanding of the T2I model.
    Specifically, we first introduce new learnable modifiers bound with classes to capture attributes of multiple concepts. 
    Then, the classes are separated and strengthened following the activation of the cross-attention operation, ensuring comprehensive and self-contained concepts.
    Additionally, we suppress the attention activation of different classes to mitigate mutual influence among concepts.
    Together, our proposed method, dubbed \textbf{DisenDiff}, can learn disentangled multiple concepts from one single image and produce novel customized images with learned concepts.
    We demonstrate that our method outperforms the current state of the art in both qualitative and quantitative evaluations.
    More importantly, our proposed techniques are compatible with LoRA and inpainting pipelines, enabling more interactive experiences.
 \end{abstract}
 
\section{Introduction}
Recently developed large-scale text-to-image models \cite{rombach2022high, saharia2022photorealistic, balaji2022ediffi, ramesh2022hierarchical} have shown unprecedented capabilities in synthesizing high-quality and diverse images based on a target text prompt. Built on these models, personalized techniques \cite{gal2022image, ruiz2023dreambooth} are further introduced to customize the models for synthesizing personal concepts with sufficient fidelity.

Given as input just a few images of the personal concepts (e.g., family, friends, pets, or individual objects), personalized text-to-image models aim to learn a new word embedding to represent a specific concept \cite{wei2023elite,tewel2023key}. However, existing methods still lack the flexibility to render all existing concepts in a given image, or only focus on a specific concept \cite{gal2023encoder,li2023blip}. Given a unique photo from a user (which could be people rarely seen together or uncommon furniture pieces), with multiple concepts occurring in the complex scene, the user naturally desires the ability to freely synthesize the concepts by composing multiple objects or focusing on only one of them. For example, two specific individuals at a beach, or alternatively, one of them in Times Square, as shown in \cref{fig:teaser}.

\begin{figure}[!t]
    \centering
    \includegraphics[width=\linewidth]{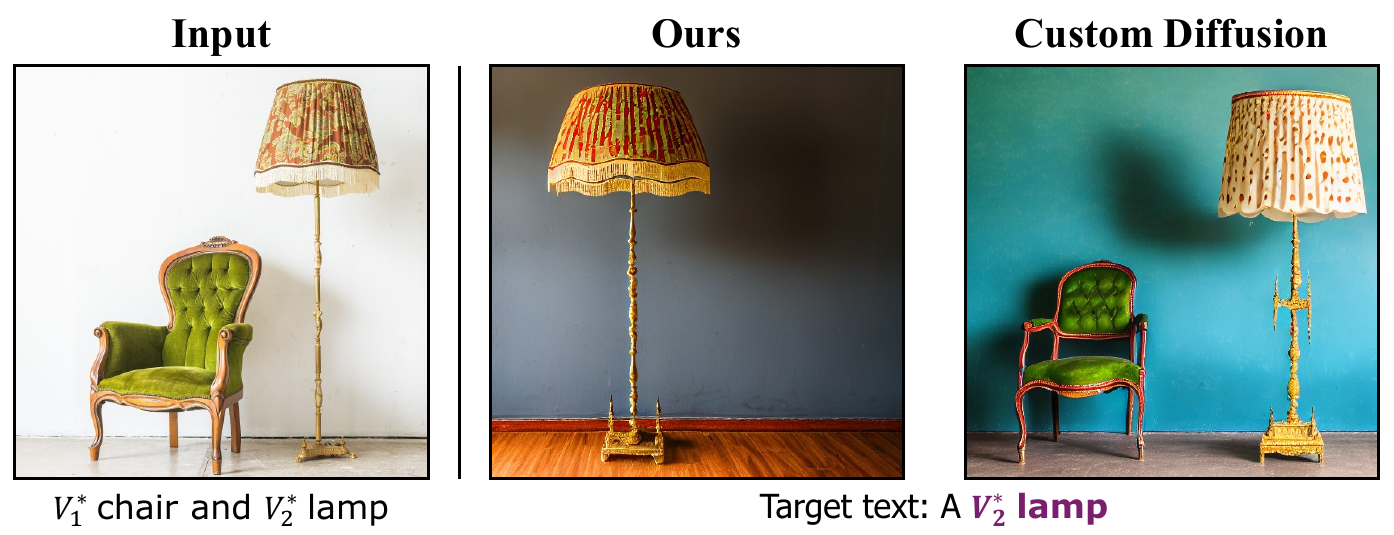}
    \caption{\textbf{Failure case of Custom Diffusion \cite{kumari2023multi}.} In the third column, we show the example encompassing two failure settings: appearance inconsistency with the input image and ambiguous object not included in the target text. In the second column, we show the result from our method.}
    \label{figure:embedding_compare}
    \vspace{-15pt}
\end{figure}

To achieve flexible renditions of the concepts, instead of using a single new word to represent one concept \cite{kumari2023multi, jia2023taming}, we employ multiple new words to represent multiple concepts. For example, considering an image containing a distinct chair and lamp (as shown in \cref{figure:embedding_compare}), we utilize the prompt ``$V_1^*$ chair and $V_2^*$ lamp'' to distinguish between them, with ``$V_1^*$'' serving as the modifier for ``chair'' and ``$V_2^*$'' as the modifier for ``lamp''. This intuitive formulation poses two key challenges. Firstly, the new word embeddings are likely to map confusing information, failing to maintain visual-fidelity to the target concepts. Secondly, with a relatively small training set (e.g., only one image), the model is prone to synthesizing multiple subjects, even when the target prompt pertains to a single concept. For example, as depicted in \cref{figure:embedding_compare}, the ideal output should exclusively feature the specified lamp when the target text is ``A $V_2^*$ lamp''. Nonetheless, the image generated by the current state-of-the-art model not only includes a lamp that doesn't match the color and texture of the input image but also involves a chair that shouldn't be present.

\begin{figure*}[!t]
    \centering
    \includegraphics[width=\linewidth]{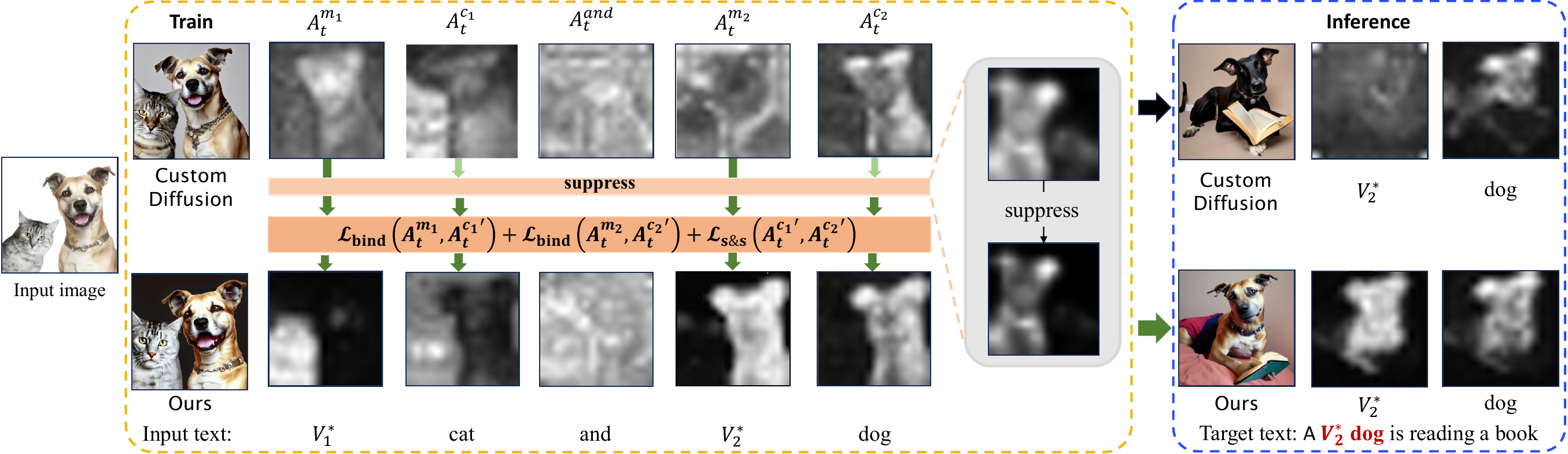}
    \caption{\textbf{Comparison of generated attention maps and images.} The first row displays the results of Custom Diffusion \cite{kumari2023multi}, while the second row shows our results. During the training stage, when we obtain accurate attention maps for important tokens (left), it leads to the ideal output during the inference stage (right), maintaining high-concept similarity with the input image.}
    \label{figure:comparsion_attn}
    \vspace{-15pt}
\end{figure*}

In this paper, we propose a novel personalized T2I model, referred to as \textit{DisenDiff} (i.e., Disentangled Diffusion), to address the above-mentioned issues. To preserve the good generalization ability in pre-trained large-scale models, we follow \cite{kumari2023multi, tewel2023key} to only update the light-weight modules ($W_K$ and $W_V$ matrices) within the cross-attention units along with new token embeddings to extend concepts. Our key insight is that current methods lack the necessary guidance for the optimization process, resulting in cluttered attention maps (as shown in \cref{figure:comparsion_attn}, the first row). Consequently, existing methods struggle to synthesize each concept effectively.

Based on the above observations, we strive to generate precise attention maps from the following two aspects. 
Building on the discovery that the attention map of the class token can roughly align with the location of the concept, then we propose a modifier-class alignment term to bind the attention map of each new modifier with its corresponding class token, correcting attention to focus on the region of the related concept. 
However, the attention maps of different class tokens often exhibit overlaps, leading to the incorrect attribute binding \cite{chefer2023attend} and mutual entanglement. To achieve effective decoupling, we introduce the separate and strengthen (s\&s) strategy to allow flexibly synthesizing each concept independently. By minimizing the overlapping regions between the attention maps of different class tokens, we can effectively mitigate the co-occurring issue when targeting at a specific concept. To further enhance the independence of concepts, we introduce a suppression technique to sharpen the boundaries of class tokens' attention maps.
Our contributions are summarized below:
\begin{itemize}
    \item We propose \textit{DisenDiff} to comprehend multiple personal concepts from only a single image. By using diverse target texts, it can render combined/independent concepts in imaginary contexts while preserving high fidelity to the input image.
    \item We employ two key constraints to attain precise attention maps for crucial tokens. The binding constraint locates new modifiers to different concepts, while the s\&s constraint decouples these concepts.
    \item We conduct experiments on various datasets and demonstrate that our method outperforms the current state of the art in quantitative and qualitative aspects. Additionally, we show the flexibility of our approach by applying it to extended tasks.
  \end{itemize}

\section{Related Work}
\myparagraph{Text-to-image generative models.} The objective of text-to-image (T2I) tasks \cite{zhu2007text, t2i2016} is to generate an image corresponding to a given textual description. Thanks to large-scale datasets \cite{schuhmann2022laion, kakaobrain2022coyo-700m} and advancements in language models \cite{kenton2019bert, radford2021learning, raffel2020T5}, T2I models have witnessed remarkable progress. While Generative adversarial networks (GANs) \cite{reed2016generative, zhang2021gant2i, kang2023gigagan, li2022stylet2i} and autoregressive (AR) transformers \cite{ramesh2021dalle, yu2022scaling, ding2022cogview2, gafni2022make} have delivered impressive results, diffusion models \cite{dhariwal2021diffusionbeatgans,ho2020ddpm} have taken the lead in T2I generation. These models employ denoising processes in image space \cite{GLIDE, saharia2022photorealistic, balaji2022ediffi, ho2022cascaded, wang2022geometric} or latent space \cite{ramesh2022hierarchical,rombach2022high, gu2022vector}, resulting in unprecedented image generation quality. However, they encounter challenges when generating specific objects, such as custom furniture, even with detailed prompts. We aim to augment these models to accurately capture the appearances of novel concepts from real-world images.

\myparagraph{Text-guided image editing.} With the surge of powerful T2I models, numerous studies have delved into enhancing the controllability of diffusion models to cater to diverse user demands. Approaches such as \cite{feng2022training, chefer2023attend, wang2023compositional} refine the cross-attention units to encompass all subject tokens, motivating the model to fully convey the semantics in the input prompt. Techniques like \cite{park2022shape, li2023gligen, chen2023trainingfree} implement region control in T2I generation by using bounding boxes and paired object labels as inputs. Additionally, \cite{zhang2023adding} and \cite{wang2022pretraining} harness pre-trained diffusion models for image-to-image translation. A substantial body of work also focuses on local or global modifications of single images using existing T2I models. Notable examples include SINE \cite{zhang2023sine} and UniTune \cite{valevski2023unitune}, which achieve image editing by fine-tuning the diffusion model. Other methods like prompt-to-prompt \cite{hertz2022prompt}, null-text inversion \cite{mokady2023null}, and \cite{patashnik2023localizing} impose constraints on latent noise during inference time without model training. While our objectives share some common ground with these methods, our primary focus is optimizing the model to seamlessly extend personalized concepts into new prompts.

\myparagraph{T2I personalization} Personalization techniques adapt diffusion models to learn new concepts from user-provided images, often relying on a small dataset of 3-5 images or even a single image. Textual Inversion \cite{gal2022image} uses pseudo-words to represent new concepts through a visual reconstruction objective. To leverage semantic priors from pre-trained models, DreamBooth \cite{ruiz2023dreambooth} utilizes a unique identifier and class name within the input text to represent new concepts. Custom-Diffusion \cite{kumari2023multi} and Perfusion \cite{tewel2023key} compose multiple new concepts by updating only the cross-attention Keys and Values along with new token embeddings. When working with a dataset containing just a single image, current methods \cite{wei2023elite, jia2023taming, gal2023encoder, li2023blip} typically begin with additional domain-specific pre-training on a large dataset before adapting to the new concept. In contrast to these methods, we aim to address the more challenging problem of acquiring multiple concepts from a single image without domain-specific pre-training.
Recently, break-a-scene \cite{avrahami2023bas} tackles the similar task using a two-phase customization method. However, its approach necessitates an extra input of an object mask, while our approach solely processes the reference image.

\begin{figure}[!t]
    \centering
    \includegraphics[width=1\linewidth]{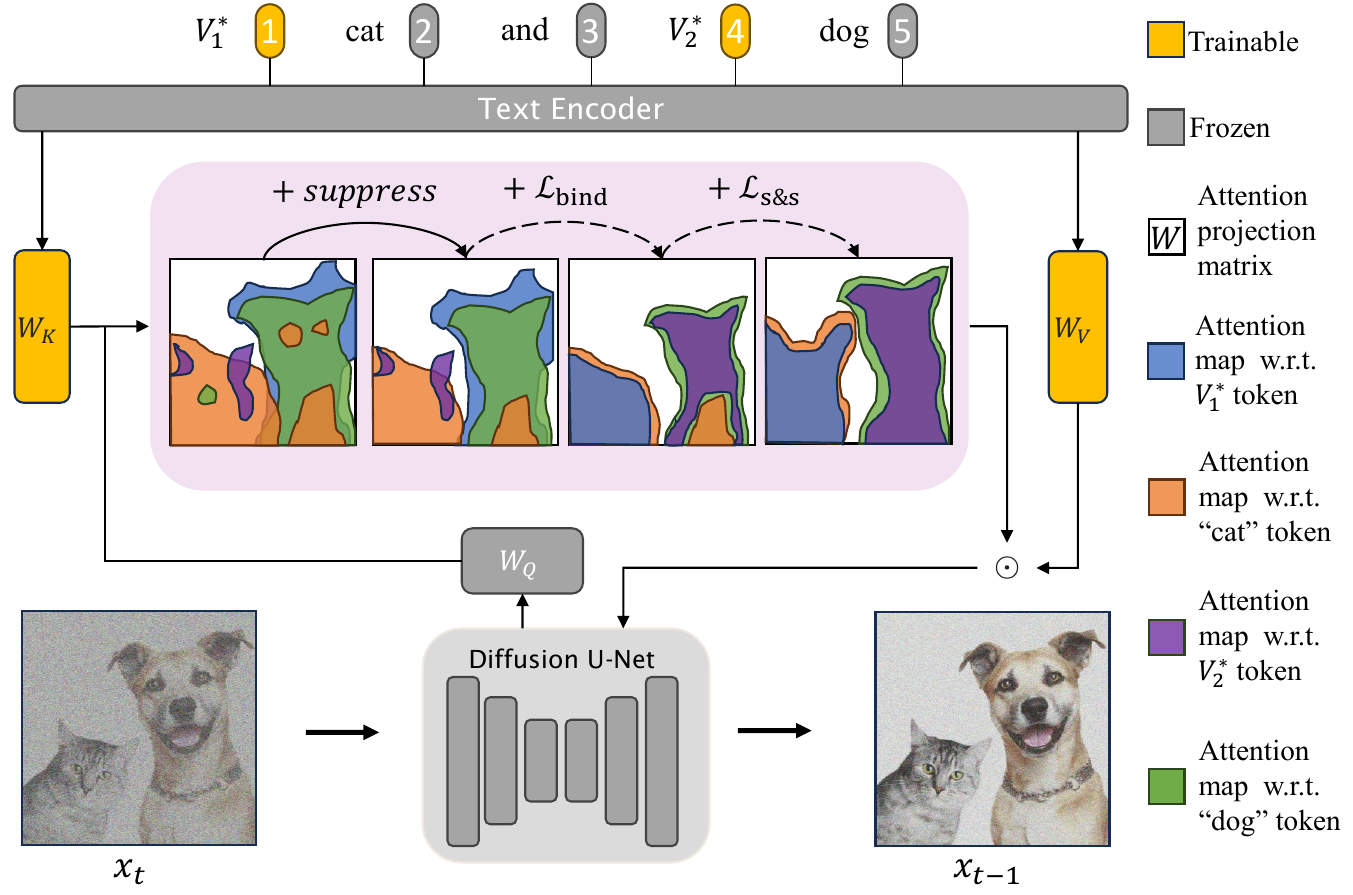}
    \caption{\textbf{Method overview.} Our method applies constraints to the cross-attention maps of crucial tokens, ensuring the accurate representation of multiple concepts. We introduce new modifiers, denoted as $V_i^*$, along with the $i$-th class name, to represent the $i$-th personalized concept. Our attention calibration mechanism mainly includes three parts: the suppression technique performs self-sharpening and filters noisy small patches, the $\mathcal{L}_{\text {bind}}$ loss steers new modifiers towards the corresponding classes, and the $\mathcal{L}_{\text {s\&s}}$ loss guarantees the independence and completeness of the learned concepts.}
    \label{figure:method}
    \vspace{-15pt}
\end{figure}

\section{Method}\label{sec:method}
Our objective is to understand multiple concepts within a single image. To this end, we propose a novel attention calibration mechanism to help generate accurate cross-attention maps in our T2I model. Firstly, the cross-attention maps are calculated as the activation responses between each word of the input text and the intermediate visual features. Then, we impose constraints on the cross-attention maps between both the modifier-class token pairs and class-class token pairs to bind the cross-attention maps of each modifier with its corresponding class (modifier-class constraint), as well as to ensure full comprehension of each class and separation between different classes (class-class constraint). To further mitigate the cross-interference issue in our T2I model, we introduce a suppression technique to obtain a sharper attention map for each class token. A schematic workflow of our method is presented in \cref{figure:method}.

\subsection{Preliminary} \label{section:preliminary}
\myparagraph{Stable Diffusion.} In our experiments, we use Stable Diffusion \cite{SD-v1-5} as our backbone model, inheriting the structure of the Latent Diffusion Model (LDM) \cite{rombach2022high}. It primarily consists of three components: a pre-trained text encoder $\tau_{\theta}$ from CLIP \cite{radford2021learning}, a VAE \cite{kingma2013auto} model $\mathcal{E}$, and a U-Net diffusion model $\epsilon_{\theta}$ trained on the latent space $z$ of the pre-trained VAE. Given the noisy latent code $z_t$ at $t$ timestep, the diffusion model predicts the random added noise $\epsilon$. The training objective of the diffusion model is formulated as follows:
\begin{equation}
    \mathbb{E}_{\mathcal{E}(x), y, \epsilon \sim \mathcal{N}(0,1), t}\left[\left\|\epsilon-\epsilon_\theta\left(z_t, t, \tau_\theta(y)\right)\right\|_2^2\right],
    \label{eq:base_loss}
\end{equation}
where $x$ denotes the input image, $y$ is the input text. Following \cite{rombach2022high}, prior knowledge in CLIP is integrated via the cross-attention mechanism.

\myparagraph{Integrating textual features via cross-attention.} Formally, the intermediate spatial representation $\phi(z_t)$ of the denoiser U-Net is mapped to a query matrix $Q=W_Q \cdot \phi(z_t)$, while text embeddings $\tau_\theta(y)$ are mapped to a key matrix $K=W_K \cdot \tau_\theta(y)$ and a value matrix $V=W_V \cdot \tau_\theta(y)$, using learnable projection matrices $W_Q$, $W_K$, and $W_V$. Then, the cross-attention maps are obtained as:
\begin{equation}
    A_t=\operatorname{Softmax}\left(\frac{Q K^T}{\sqrt{d}}\right),
\end{equation}
where $d$ is the projection dimension of keys $K$ and queries $Q$. Here, $A_t \in \mathbb{R}^{r \times r \times N}$, $r$ is the spatial dimension of the $\phi(z_t)$, and $N$ is the number of input tokens.
The updated spatial representations integrating text priors are then obtained as $\phi(z_t) = A_tV$, as illustrated in \cref{figure:method}.

\myparagraph{Text encoding.} Generally, during training of a T2I system, a suitable text prompt is required in addition to the selected single image.
In this paper, we adopt a manner similar to \cite{ruiz2023dreambooth}, incorporating new modifiers and the classes to be modified into the input text. For example, if the target image contains a cat and a dog, the text prompt would be ``$V_1^*$ cat and $V_2^*$ dog''. The modifier tokens ``$V_i^*$'' are initialized with rare vocabulary. Given only a single training image, the T2I model will likely lack the diversity of generation, known as the language drift \cite{lee2019countering,lu2020countering} problem. Using our text prompt, we can easily select regularized images with the same caption to mitigate the issue of language drift, enabling our model to generate a variety of cats and dogs (not limited to the ones present in the target images, as shown in \cref{figure:qualitative_comparsion}, left of the second row). 

Current methods are prone to overfitting when the training data only consists of a single image, resulting in ambiguous attention maps for each token (as shown in the first row of \cref{figure:comparsion_attn}). As demonstrated in P2P \cite{hertz2022prompt}, the spatial layout and geometry of the generated images depend on the cross-attention maps. Therefore, our primary focus is to optimize the model to produce accurate cross-attention maps, elaborated in the following part.

\subsection{Coherent binding of modifiers with classes}

Based on the cross-attention maps ($A_t$) obtained by a previous method (shown in \cref{figure:comparsion_attn}, the first row), we can observe that while $A_t$ of new modifiers are chaotic ($A_t^{m_1}$ and $A_t^{m_2}$), cross-attention of class tokens can roughly capture the semantic boundaries ($A_t^{c_1}$ and $A_t^{c_2}$). We attribute it to the fact that the majority of parameters in the T2I models are frozen, preserving the category information of class tokens. To aid the new modifiers in understanding their responsibilities, we define the constraint to bind the cross-attention maps of modifiers with their corresponding class tokens as
\begin{equation}
    \mathcal{L}_{\text {bind}}\left(A_t^{m_i}, A_t^{c_i}\right)=1-\frac{A_t^{m_i} \cap A_t^{c_i}}{A_t^{m_i} \cup A_t^{c_i}},
\end{equation}
where $A_t^{m_i}$ and $A_t^{c_i}$ represent the attention map of the $i$-th modifier and the $i$-th class at $t$ timestep, respectively. The $\mathcal{L}_{\text {bind}}$ loss is formulated to reduce the intersection over union (IoU) \cite{yu2016unitbox} between these two attention maps, encouraging a close alignment between the activations of the modifiers and the class tokens. 
To prevent substantial influence on $A_t^{c_i}$, we detach its gradient during the loss computation.

Nonetheless, there are two potential issues when we directly apply this constraint. Given that $A_t$ is the result of the Softmax operation (i.e., ${\textstyle \sum_{i=1}^{N}A_t^i(h,w)}=1$, where $A_t^i(h,w)$ denotes the activation of the $i$-th token at pixel $(h,w)$), input tokens would contend for attention at the same position. Consequently, a precise pixel-to-pixel correspondence between $A_t^{m_i}$ and $A_t^{c_i}$ can not be established. Furthermore, our intention is for the activations of $A_t^{m_i}$ to fully encompass the corresponding object, thereby capturing all its attributes comprehensively. However, as depicted in \cref{figure:comparsion_attn}, it is evident that within the object region, certain activations of $A_t^{c_2}$ exhibit high values, while others appear considerably lower. This poses a challenge for the attention $A_t^{m_i}$ to sustain a comprehensive focus on the object. To address these challenges, we employ a Gaussian filter on $A_t$, which leads to the generation of smooth attention maps referred to as $G(A_t)$. This smoothing process helps to alleviate the pixel-wise competition among tokens and facilitates more comprehensive attention to the object. Consequently, by using the loss function $L_{\text {bind}}\left(G(A_t^{m_i}), G(A_t^{c_i})\right)$, we encourage $A_t^{m_i}$ to have coherent attention areas with $A_t^{c_i}$, while achieving a broader coverage of the object, without the need for precise point-to-point binding. 
For simplicity, in the subsequent sections of this paper, unless explicitly specified otherwise, we apply a Gaussian filter to $A_t$.

\subsection{Separating and strengthening attention maps for multiple classes} \label{sec:ss}
\begin{figure*}[!t]
    \centering
    \includegraphics[width=\linewidth]{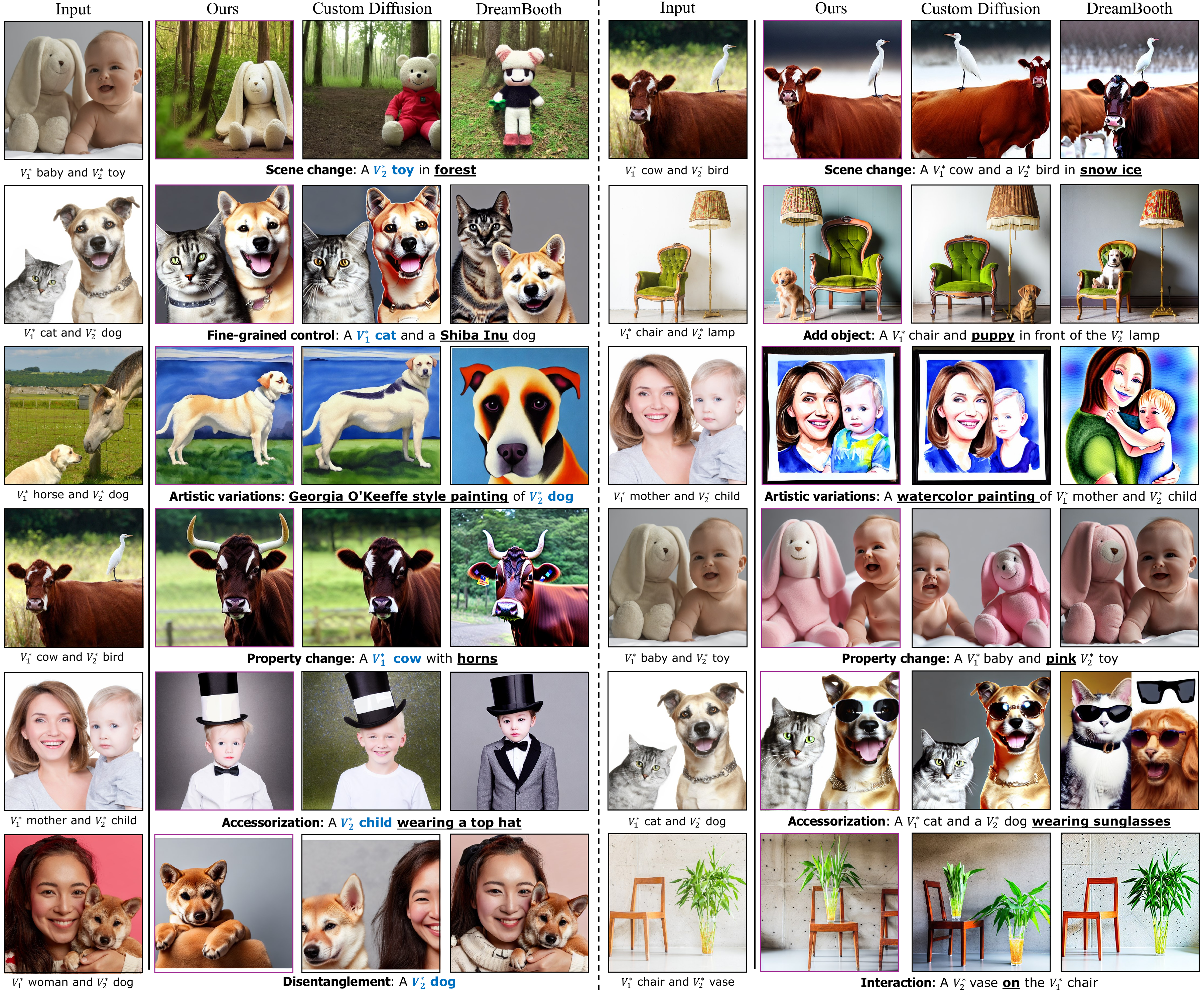}
    \caption{\textbf{Qualitative results of independent (left) and combined (right) concepts.} The target prompt in each row represents a distinct context including learned concepts. Our method shows the highest visual similarity to the input image compared to Custom Diffusion and DreamBooth (especially in the first row, the results containing the specific toy) while preserving robust editability. Additionally, we show the ability to address the language drift issue and the disentanglement capability on the left of the second and last row, respectively. } 
    \label{figure:qualitative_comparsion}
    \vspace{-15pt}
\end{figure*}
Given a single image as the training set, it's inevitable for one class token to attend to multiple concepts simultaneously. For instance, in the first row of \cref{figure:comparsion_attn}, specifically in $A_t^{c_1}$, the attention dedicated to the ``cat'' token is not solely limited to the ``cat'' concept. It also exhibits some degree of attention towards the ``dog'' concept. Thus, $A_t^{m_1}$ incorporates attributes associated with the ``dog'' concept due to its binding with $A_t^{c_1}$. To ensure independent editing of concepts without interference, it is necessary to separate the attention regions of different objects (i.e., $A_t^{c_i}$ and $A_t^{c_j}$). A straightforward approach is to minimize the overlap between attention maps of different object tokens as
\begin{equation}
    \mathcal{L}_{\text {separate}}\left(A_t^{c_i}, A_t^{c_j}\right)=A_t^{c_i} \cap A_t^{c_j}.
\end{equation}
The utilization of $\mathcal{L}_{\text {separate}}$ effectively prevents the activations of class tokens from overlapping. However, it may come with a side effect of reducing the area of $A_t^{c_i}$, potentially leading to a loss of identity for the corresponding class, which can be found in the supplement. To simultaneously minimize the overlap among attention maps and preserve the class identity, we design the following constraint, 

\begin{equation}
    \mathcal{L}_{\text {s\&s}}\left(A_t^{c_i}, A_t^{c_j}\right)=\frac{A_t^{c_i} \cap A_t^{c_j}}{A_t^{c_i} \cup A_t^{c_j}},
\end{equation}
where ``s\&s'' stands for ``separate and strengthen.'' The $\mathcal{L}_{\text {s\&s}}$ loss strikes a balance between avoiding overlap with other objects and ensuring comprehensive coverage of the target object, thus improving the accuracy and fidelity of the attention mechanism.

\myparagraph{Suppression.}
The utilization of the $\mathcal{L}_{\text{s\&s}}$ loss can potentially lead to another issue where the attention map $A_t^{c_1}$ captures a significant portion of the activations, while $A_t^{c_2}$ exhibits very few activations. This imbalance in activation distribution between different class tokens can result in an uneven emphasis on certain classes. To address it, we introduce a suppression mechanism. Specifically, before computing the $\mathcal{L}_{\text {s\&s}}$, we apply an element-wise multiplication operation to $A_t^{c_i}$ (i.e., $f_{m}(A_t^{c_i})=A_t^{c_i}\odot A_t^{c_i}$). Given that activations fall within the range of $[0,1]$, $f_{m}(A_t^{c_i})$ filters out activations that are less important for the class. As a result, the loss $L_{\text {s\&s}}(f_{m}(A_t^{c_i}), f_{m}(A_t^{c_j}))$ is designed to separate and strengthen their attentions, preventing encroachment upon other classes from within its own boundaries. Additionally, $A_t^{m_i}$ can be bound with a more distinct $A_t^{c_i}$. 

In summary, the total training loss is formulated as:
\begin{equation}
    \label{eq:total_loss}
    \begin{aligned}
        \mathcal{L} &= \mathcal{L}_{\text {base}} + \sum_{i=1}^{\mathcal{S}}\mathcal{L}_{\text {bind}}\left(G(A_t^{m_i}), f_{m}(G(A_t^{c_i}))\right)  \\
        &+ \sum_{i=1}^{\mathcal{S}}\sum_{j=i+1}^{\mathcal{S}}\mathcal{L}_{\text {s\&s}}\left(f_{m}(G(A_t^{c_i})), f_{m}(G(A_t^{c_j}))\right),
    \end{aligned}
\end{equation}
where $\mathcal{S}$ is the number of classes in the input image, and $\mathcal{L}_{\text {base}}$ is the base loss of the T2I model in \cref{eq:base_loss}.
$\mathcal{L}_{\text {s\&s}}$ is responsible for refining the attention maps related to class tokens, while $\mathcal{L}_{\text {bind}}$ is responsible for constraining new modifier tokens to acquire correct attributes.
The auxiliary functions $G(\cdot)$ and $f_{m}(\cdot)$ facilitate the optimization process.
The synergy among these constraints results in the generation of precise and interpretable attention maps for input tokens, shown in the second row of \cref{figure:comparsion_attn}.
\section{Experiments}

\subsection{Experimental Settings} \label{section:settings}
\myparagraph{Datasets.} We conducted experiments on ten datasets spanning a large range of categories including people, animals, furniture, and people with pets/toys. Please note, instead of concentrating only on one concept, our datasets contain \textit{two distinct concepts} within each image.  During the inference phase, we test 30 different prompts for each image: 10  for combined concepts, 10 specifically targeting the first concept, and 10 focusing on the second concept. 

\begin{figure*}[!t]
    \centering
    \includegraphics[width=\linewidth]{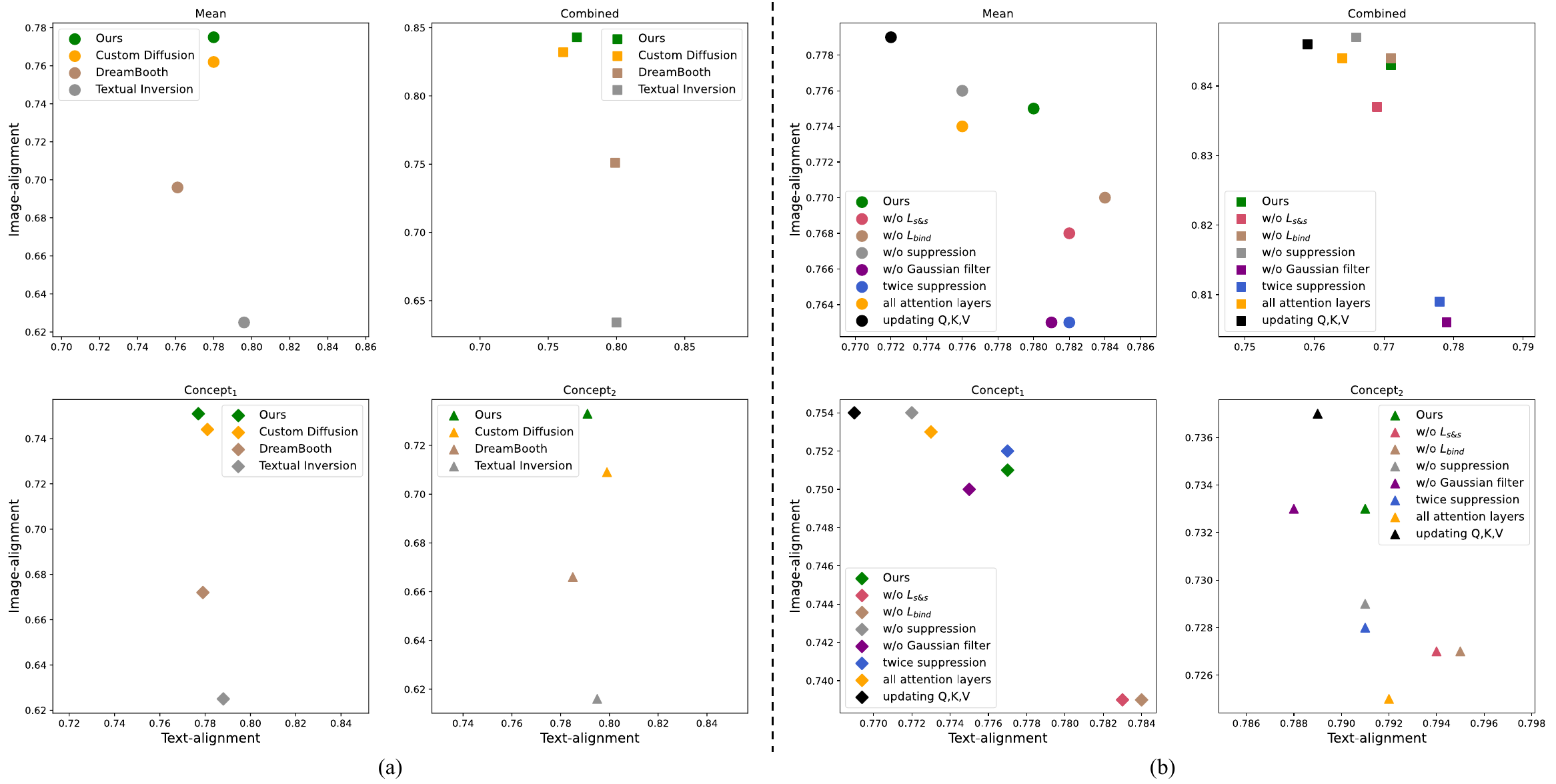}
    \caption{\textbf{Quantitative evaluation results.} (a) Compared to state-of-the-art methods, our approach (green) achieves the highest image-alignment score, particularly noticeable in Concept$_2$, while maintaining a text-alignment score similar to that of other methods. (b) Ablation study results. Our full method (green) strikes the best balance between reconstruction and editability.}
    \label{figure:main_comparsion}
    \vspace{-15pt}
\end{figure*}

\myparagraph{Compared methods.} We compare with three personalized T2I methods, which all utilize new word embeddings to represent novel concepts. (1) Textual Inversion (TI): In TI, only the new token embedding representing the novel concept is updated, while the other parameters remain frozen. (2) DreamBooth (DB): DB updates all layers of the T2I model to maintain visual fidelity and employs a prior preservation loss to mitigate language drift. (3) Custom-Diffusion (CD): CD updates the most relevant weights related to the input textual features, including $W_K$ and $W_V$ within the cross-attention units, as well as the new token embedding. The implementation details are provided in the supplement.

\myparagraph{Evaluation metrics.} The synthetic images should faithfully capture the visual characteristics of the input image while accurately conveying all elements of the target text. We employ two key metrics: (1) The image-alignment metric evaluates the reconstruction of concepts, which measures the pairwise CLIP-space cosine similarity \cite{gal2022image} between the generated images and the corresponding real images. (2) The text-alignment metric assesses the editing effectiveness of the fine-tuned model by calculating the text-image similarity between the generated images and the provided prompts using CLIP \cite{hessel2021clipscore}. Notably, these two indicators often conflict with each other \cite{tewel2023key}. For each concept, we synthesize 16 samples per prompt, using 50 DDIM steps and a guidance scale of 6. For comparison, we provide scores for combined concepts, the first concept, the second concept, and their average (referred to as Combined, Concept$_1$, Concept$_2$, and Mean in \cref{figure:main_comparsion}). For instance, if the training image caption is ``$V_1^*$ cat and $V_2^*$ dog'', the test prompts of the Combined, Concept$_1$ and Concept$_2$ settings are ``$V_1^*$ cat and $V_2^*$ dog in a garden'', ``$V_1^*$ cat wearing a hat'', ``A pink $V_2^*$ dog'', respectively. When testing on independent concepts, we calculate the image-alignment metric between the synthesized images and the segmented image containing only the corresponding subject.

\myparagraph{Implementation details.} We fine-tune the Stable Diffusion \cite{SD-v1-5} model for 250 steps, with a batch size of 8 and a learning rate of $8 \times 10^{-5}$. Similar to \cite{kumari2023multi}, we employ clip-retrieval \cite{beaumont-2022-clip-retrieval} to select $200$ samples from LAION-5B \cite{schuhmann2022laion} dataset as regularization images. Captions of these selected images exhibit a similarity of over 0.85 in the CLIP textual embedding space with the input text. Meanwhile, we use the data augmentation in \cite{kumari2023multi}. In our experiments, we apply the proposed cross-attention calibration to the $16 \times 16$ attention units, which have been shown to contain the most semantic information \cite{hertz2022prompt}.

\subsection{Comparison Results} \label{section:comparsion_results}
\myparagraph{Quantitative comparisons.} \cref{figure:main_comparsion}a illustrates the results averaged across ten datasets. As shown, we outperform all the compared methods, especially on the image-alignment scores. Specifically, despite Textual Inversion (TI) achieving the highest text-alignment score, it has the lowest image-alignment score, indicating its struggle to maintain the appearance of concepts. DreamBooth (DB) outperforms TI in image-alignment score but falls significantly short compared to our approach in both metrics. Custom Diffusion (CD) maintains a better balance between the two metrics and competes with ours in combined concepts scores and Concept$_1$ scores. However, there is a noticeable performance gap in the scores for Concept$_2$. In summary, we achieve the highest image fidelity while maintaining strong text editing effectiveness. Detailed results for each dataset can be found in the supplement.

\myparagraph{Qualitative comparisons.} We visually demonstrate the favorable outcomes in \cref{figure:qualitative_comparsion}. Concretely, we design diverse target prompts to assess the learned independent concepts and combined concepts in different editing scenarios, including scene changes, object addition, style transfer, property change, accessory addition, interactions between multiple concepts, concept decoupling, and the ability to address the language drift (e.g., generating a specific cat consistent with the input and a dog with a breed distinct from the one present in the input). As shown in \cref{figure:qualitative_comparsion}, images synthesized by DB either lack key attributes of the concepts or suffer from severe overfitting to the input image. With most of its parameters frozen, CD improves editability and reconstruction compared to DB. However, it still struggles to preserve concepts' appearances or decouple from the input image, especially as shown in the first and last rows of \cref{figure:qualitative_comparsion}. By incorporating cross-attention calibration, our method achieves high visual fidelity and maintains effective cross-concept disentanglement during T2I generation. For the sake of space efficiency, additional results including Textual Inversion are provided in the supplement.

\subsection{Ablation Studies} 

We conduct ablation studies to show the effectiveness of each component and analyze the influence of different design choices, adopting the same setup described in \cref{section:settings}.

To assess the necessity of each component, we set up the following experiment settings: (1) Removing the $\mathcal{L}_{\text {s\&s}}$ loss, (2) removing the $\mathcal{L}_{\text {bind}}$ loss, (3) removing the suppression strategy, (4) removing the Gaussian filter,  (5) applying twice suppression (in contrast to one-time). Detailed results are presented in \cref{figure:main_comparsion}b. As shown, our full model achieves a balanced performance between visual fidelity and editing effectiveness for both combined and independent concepts. Removing either the $\mathcal{L}_{\text {bind}}$ or $\mathcal{L}_{\text {s\&s}}$ loss results in a significant decrease in image-alignment for both Concept$_1$ and Concept$_2$. Similarly, the removal of the Gaussian filter leads to a notable reduction in image-alignment for combined concepts. No suppression significantly harms image-alignment for Concept$_2$, confirming the benefits of sharper boundaries in $A_t^{s_i}$ for understanding multiple concepts (as explained in \cref{sec:ss}). Meanwhile, this also leads to lower text-alignment for both Concept$_1$ and Concept$_2$. Furthermore, applying twice suppression has detrimental effects on image-alignment as it filters out important information.

\begin{figure}[!t]
    \centering
    \includegraphics[width=\linewidth]{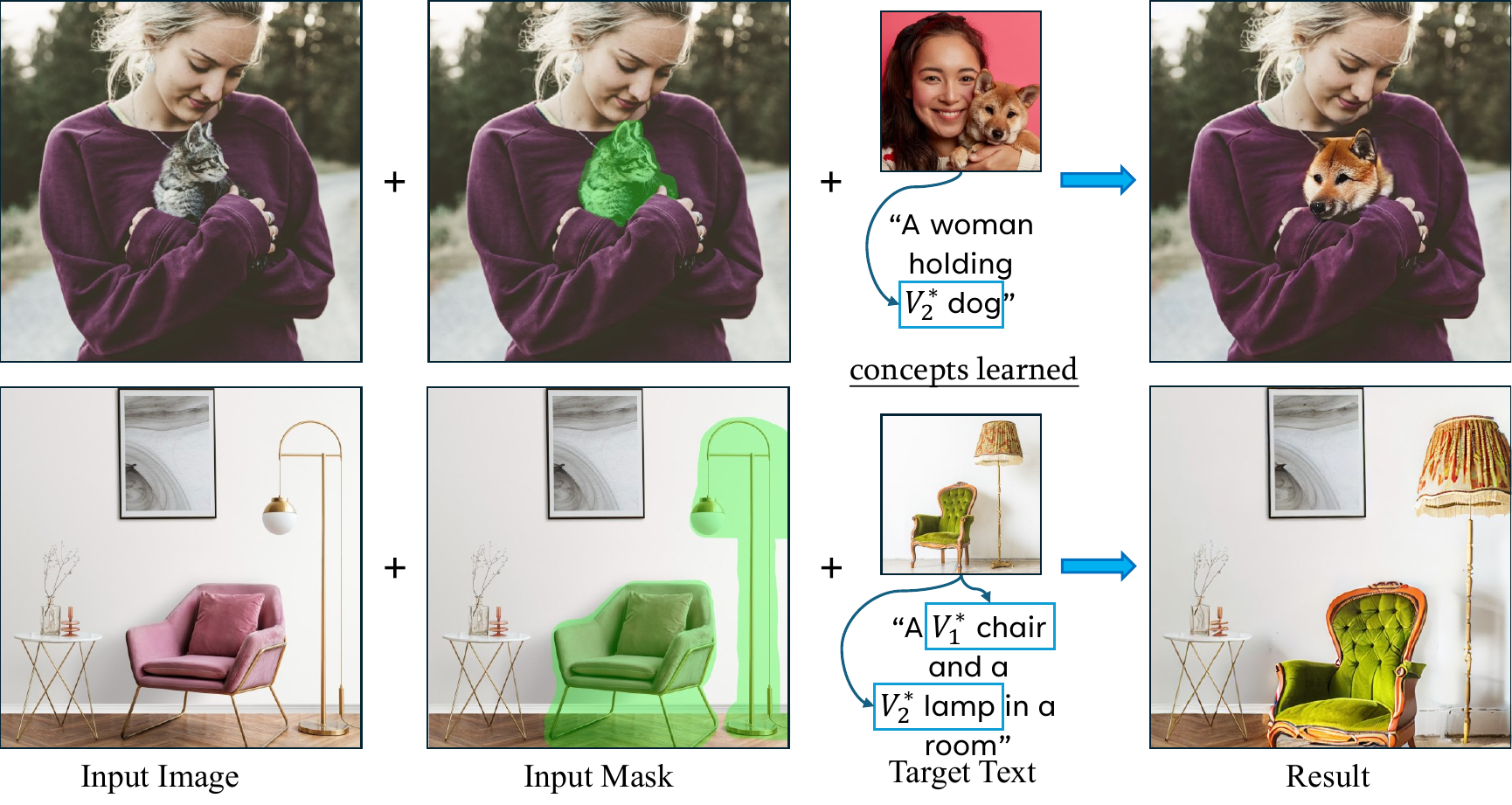}
    \caption{\textbf{Applications in image inpainting.} Given an input image and its corresponding mask, our method can seamlessly inpaint the learned concepts into the masked region.}
    \label{figure:inpainting}
    \vspace{-13pt}
\end{figure}
\begin{figure}[!t]
    \centering
    \includegraphics[width=\linewidth]{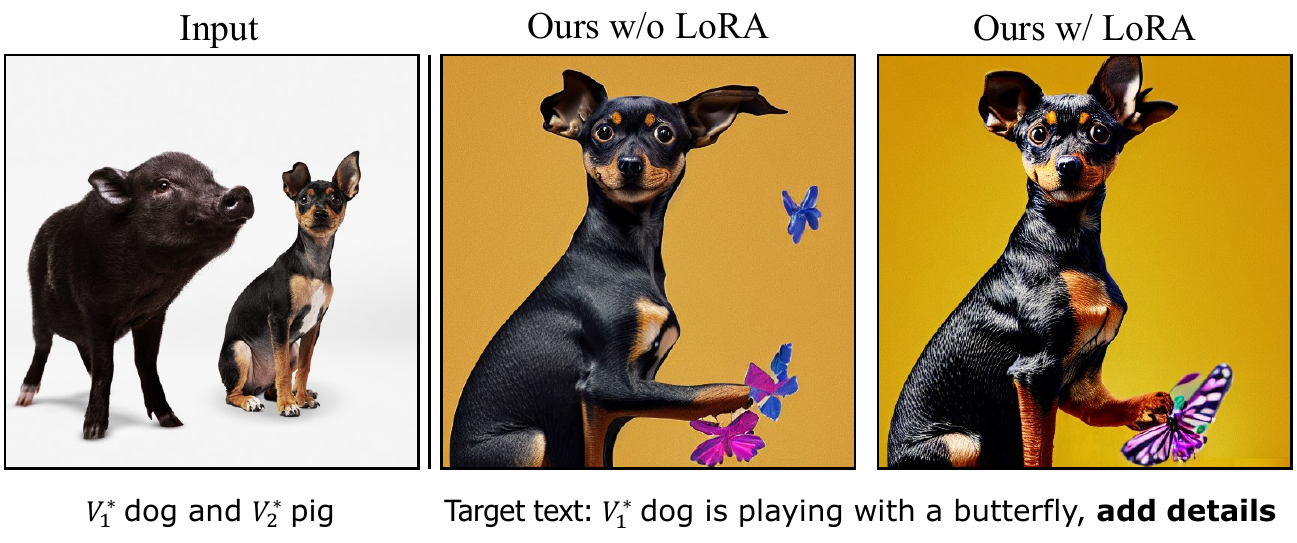}
    \caption{\textbf{Integrating with LoRA \cite{hu2021lora}.} Our method can incorporate the LoRA parameters to fully convey the semantics (e.g., enhancing texture details).}
    \label{figure:LoRA}
    \vspace{-15pt}
\end{figure}

On the other hand, there are two design choices worth considering. As indicated in \cite{tang2022daam}, averaging all scales of attention layers, instead of just using the $16 \times 16$ scale, could potentially yield improved attribution maps for each input word. Therefore, we explore (1) impose constraints on the average of all scales attention layers. Additionally, we investigate releasing more parameters, specifically (2) updating the $W_Q$, $W_K$, and $W_V$ matrices within the cross-attention units (in contrast to our approach, which only updates the $W_K$ and $W_V$). As depicted in  \cref{figure:main_comparsion}b, operating on all scales of attention layers resulted in the model's inability to reconstruct Concept$_2$. Updating $W_Q$, $W_K$, and $W_V$ does help the model remember the appearances of concepts but leads to a significant decrease in text-alignment. This suggests that updating more parameters does not preserve the good features of the pre-trained model.

\subsection{Applications}

\myparagraph{Personalized concept inpainting.} With any image and its corresponding mask, our method can seamlessly integrate learned concepts into the masked region while preserving the rest of the image, as shown in \cref{figure:inpainting}. Users can effortlessly perform inpainting by simply modifying the text prompt, thanks to our method's conversion of concepts into new word embeddings.

\myparagraph{Compatible with LoRA \cite{hu2021lora}.} LoRA techniques, actively discussed in the community, such as CivitAI \cite{CivitAI}, have gained popularity for enhancing specific capabilities of T2I models, such as improving the ability to refine images. LoRA adds small, trainable parameters to the frozen T2I models for fine-tuning, and our method is orthogonal with it. Therefore, we combine the LoRA with our trained model to unlock a wider range of applications, as shown in \cref{figure:LoRA}. This combination is akin to domain-specific pre-training on a large dataset before personalization \cite{li2023blip, gal2023encoder}, with the added benefit of having access to a wealth of readily available LoRA parameters in the community.

\begin{figure}[!t]
    \centering
    \includegraphics[width=\linewidth]{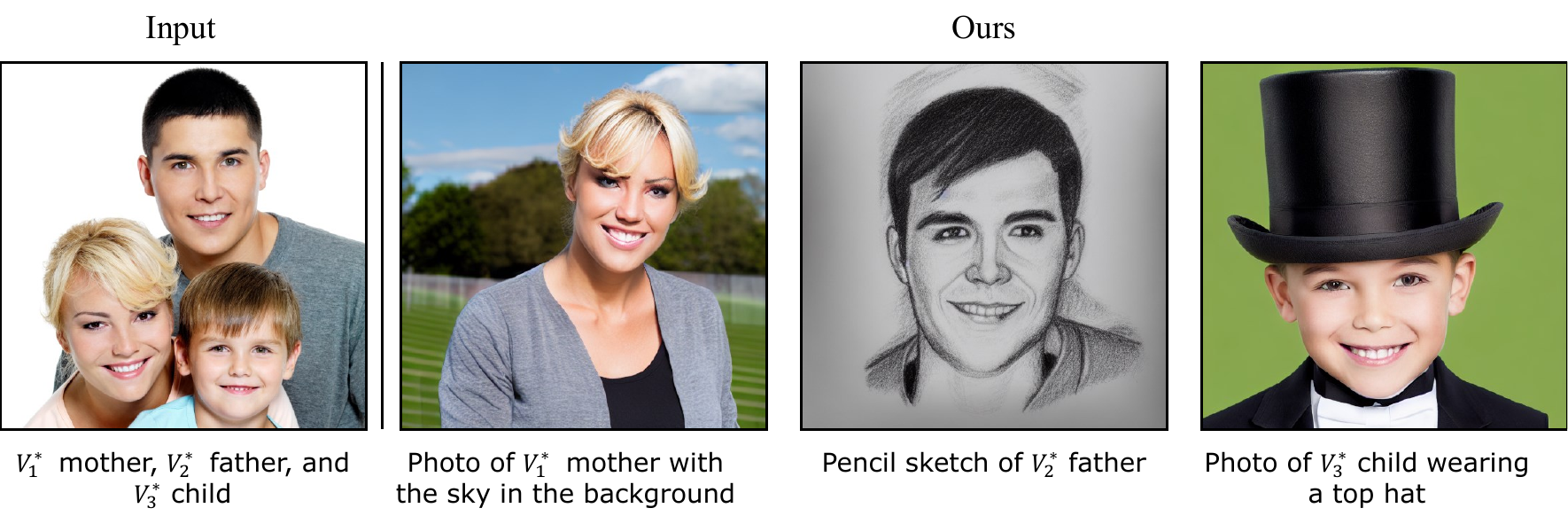}
    \caption{\textbf{Applications in extending three concepts.} Enabling edits on three concepts within a single image. }
    \label{figure:three_concepts}
    \vspace{-15pt}
\end{figure}

\myparagraph{Extending to three concepts.} We explore the application of our method to the more challenging task of capturing three concepts from a single image, as shown in \cref{figure:three_concepts}. In this scenario, we employ the $\mathcal{L}_{\text {s\&s}}$ loss for each pair of the three class tokens to disentangle these concepts.
\section{Conclusions and Limitations} 
We propose the \textit{DisenDiff} to mimic multiple concepts from a single image. We introduce constraints on the cross-attention units to attain precise attention maps for crucial tokens, mitigating the overfitting to the single image and accurately capturing concept appearances. Consequently, our method enables diverse edits involving combined or independent concepts while enhancing the visual similarity between the synthesized images and the input image. Furthermore, we show the flexibility of our method by evaluating several applications.

\myparagraph{Limitations.} Disentangling fine-grained categories becomes notably challenging when two subjects from the same category co-exist in a single image, such as Golden Retriever and Border Collie dogs. Additionally, while our method can handle images with three concepts, its performance degrades considerably. This can be attributed to the limitations of existing T2I models in such scenarios, as well as the need for algorithm adjustments to address these specific challenges. We believe that there is considerable room to enhance the performance in these complex tasks.

\myparagraph{Acknowledgment.} This work is supported by Shanghai Science and Technology Program "Federated based cross-domain and cross-task incremental learning" under Grant No. 21511100800, Natural Science Foundation of China under Grant No. 62076094 and No. 62201341.


{
    \small
    \bibliographystyle{ieeenat_fullname}
    \bibliography{main}
}

\clearpage
\setcounter{page}{1}
\maketitlesupplementary

\section{Experiments}

\begin{figure*}[!t]
    \centering
    \includegraphics[width=\linewidth]{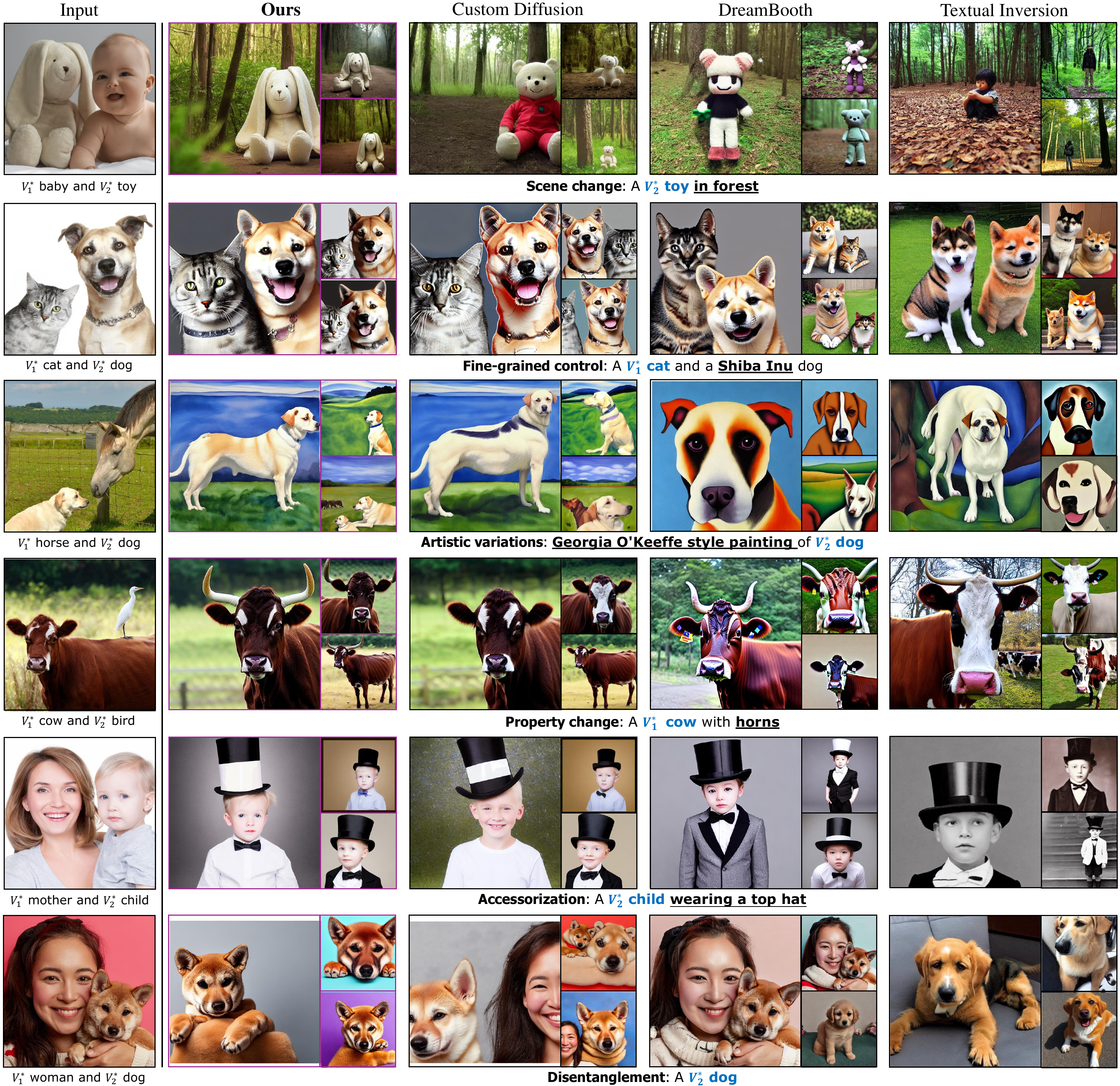}
    \caption{\textbf{Qualitative comparison on independent concepts including Textual Inversion.} } 
    \label{figure:qual_single_subject_w_ti}
    \vspace{-15pt}
\end{figure*}

\begin{figure*}[!t]
    \centering
    \includegraphics[width=\linewidth]{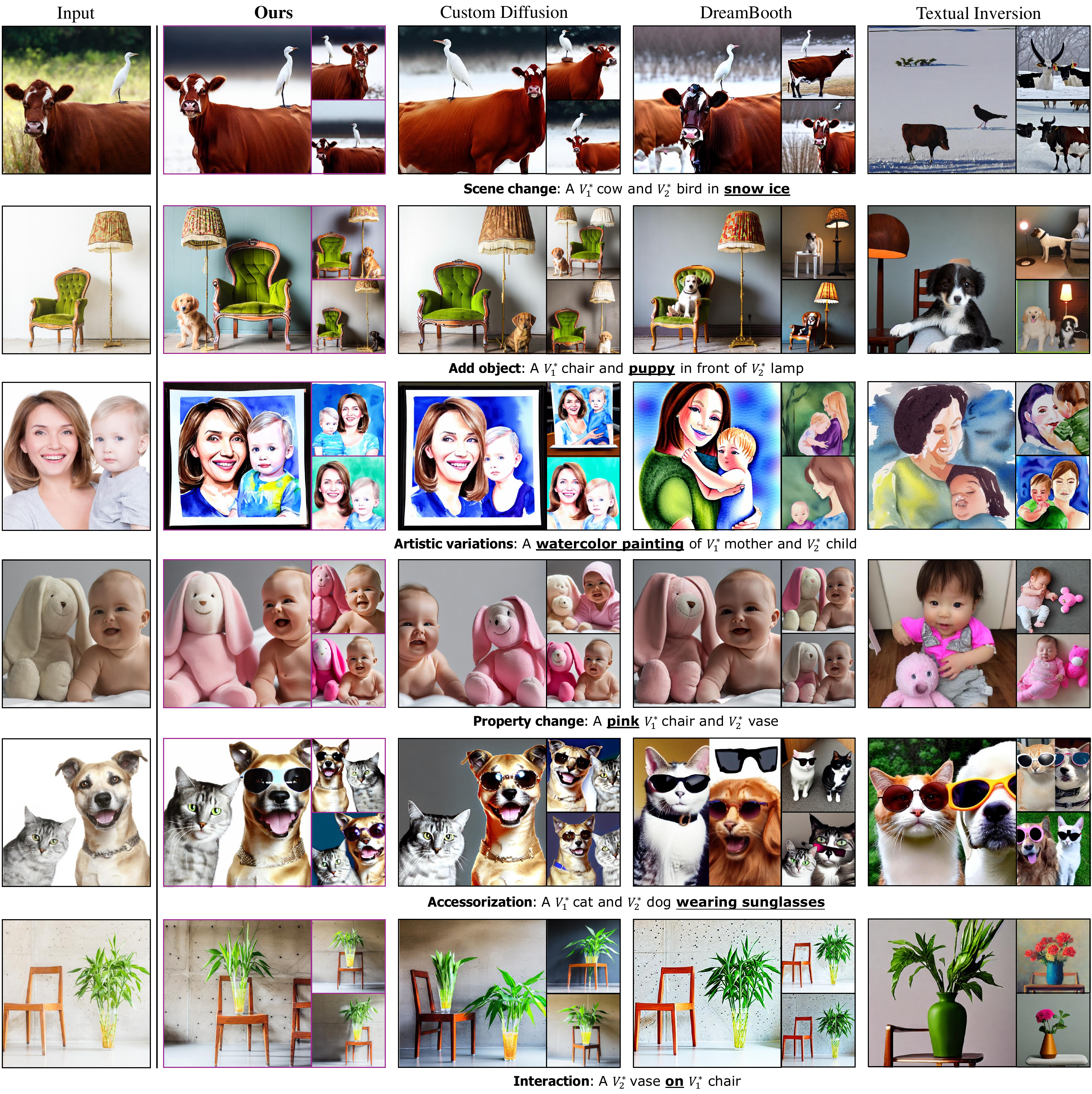}
    \caption{\textbf{Qualitative comparison on combined concepts including Textual Inversion.} } 
    \label{figure:qual_combined_subject_w_ti}
    \vspace{-15pt}
\end{figure*}

\myparagraph{Additional qualitative results.} Further comparisons, including Textual Inversion (TI) \cite{gal2022image}, are illustrated in Figure \ref{figure:qual_single_subject_w_ti} (independent concepts) and Figure \ref{figure:qual_combined_subject_w_ti} (combined concepts). Evidently, the concepts synthesized by TI differ significantly from the input image, affirming the quantitative analysis in \cref{section:comparsion_results}.

\myparagraph{Detailed quantitative results on ten datasets.} As shown in \cref{table:quantitative_results_each_dataset}, our method consistently attains the highest image-alignment across most datasets while maintaining favorable text-alignment compared to the three baselines.

\begin{figure*}[!t]
    \centering
    \includegraphics[width=\linewidth]{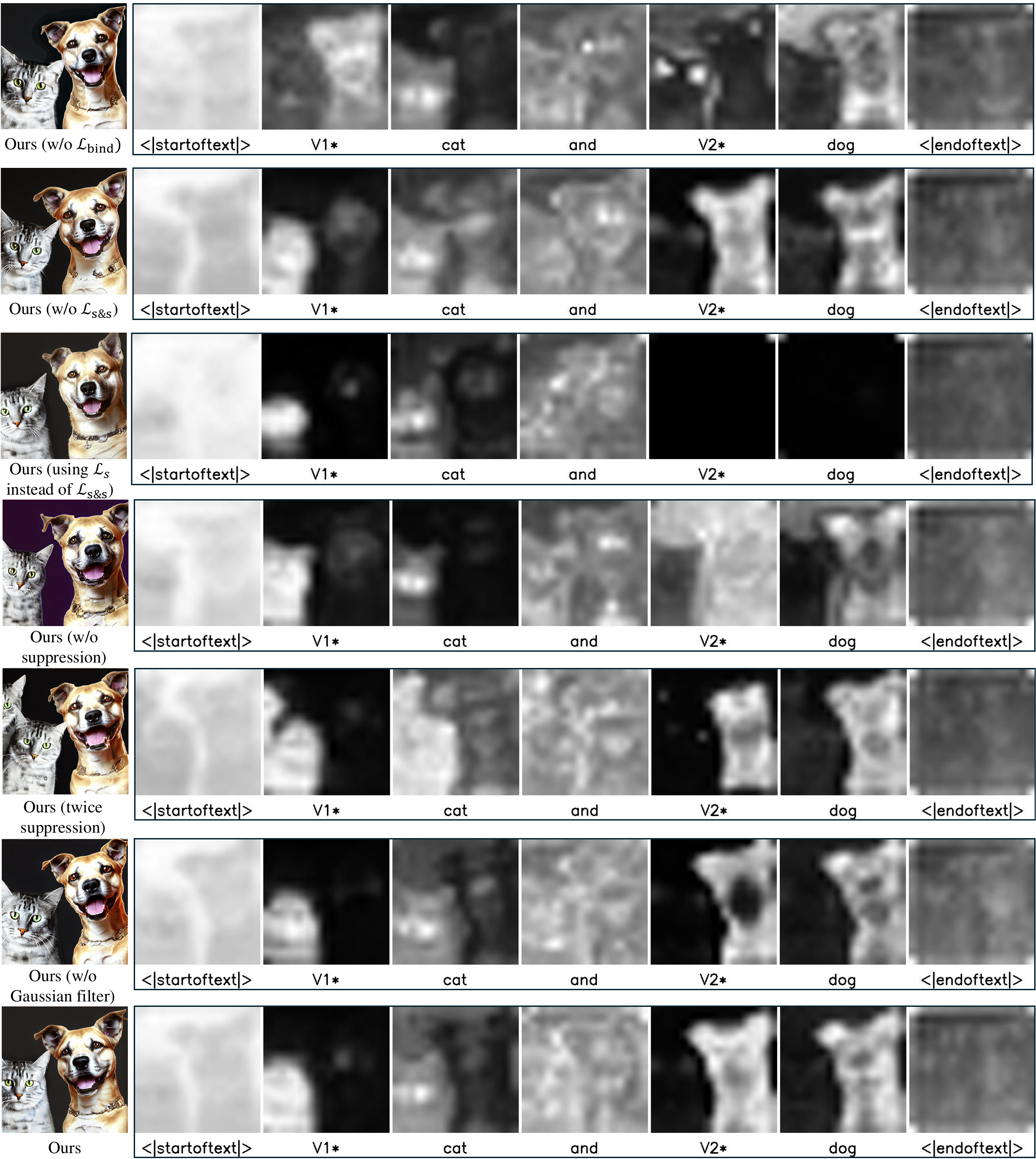}
    \caption{\textbf{Attention map visualization of ablations. } Each row represents the generated image and attention maps for all input tokens by ablation methods.} 
    \label{figure:attn_ablations}
    \vspace{-15pt}
\end{figure*}

\myparagraph{Attention map visualization of ablation studies.} The attention maps for the component ablations are presented in \cref{figure:attn_ablations}, encompassing the following scenarios: (1) Removing the $\mathcal{L}_{\text {bind}}$ loss, (2) removing the $\mathcal{L}_{\text {s\&s}}$ loss, (3) using $\mathcal{L}_s$ (i.e., $\mathcal{L}_{\text {separate}}$ in \cref{sec:ss}) instead of $\mathcal{L}_{\text {s\&s}}$, (4) removing the suppression strategy, (5) applying twice suppression, (6) removing the Gaussian filter. Observing \cref{figure:attn_ablations} reveals the following insights: (1) Without $\mathcal{L}_{\text {bind}}$, new modifiers tend to focus on incorrect classes or vague regions; (2) Absence of $\mathcal{L}_{\text {s\&s}}$ results in interdependence among learned class tokens, especially the ``cat'' token; (3) Sole reliance on $\mathcal{L}_s$ leads to tiny activation areas for crucial tokens; (4) Removal of the suppression strategy introduces unnecessary activations for new modifiers, apart from their corresponding class regions; (5) Applying twice suppression causes the loss of vital information for new modifiers, (e.g., the attention of $V_2^*$ is obviously smaller than the ``dog''); (6) The absence of the Gaussian filter may cause new modifiers to lack specific attributes related to the concepts, such as the attention on the mouth part for $V_2^*$ in the specific dog instance. In summary, our full method generates independent and comprehensive attention maps for crucial tokens.

\section{Implementation and Experiment Details}
\begin{figure*}[h]
    \centering
    \includegraphics[width=\linewidth]{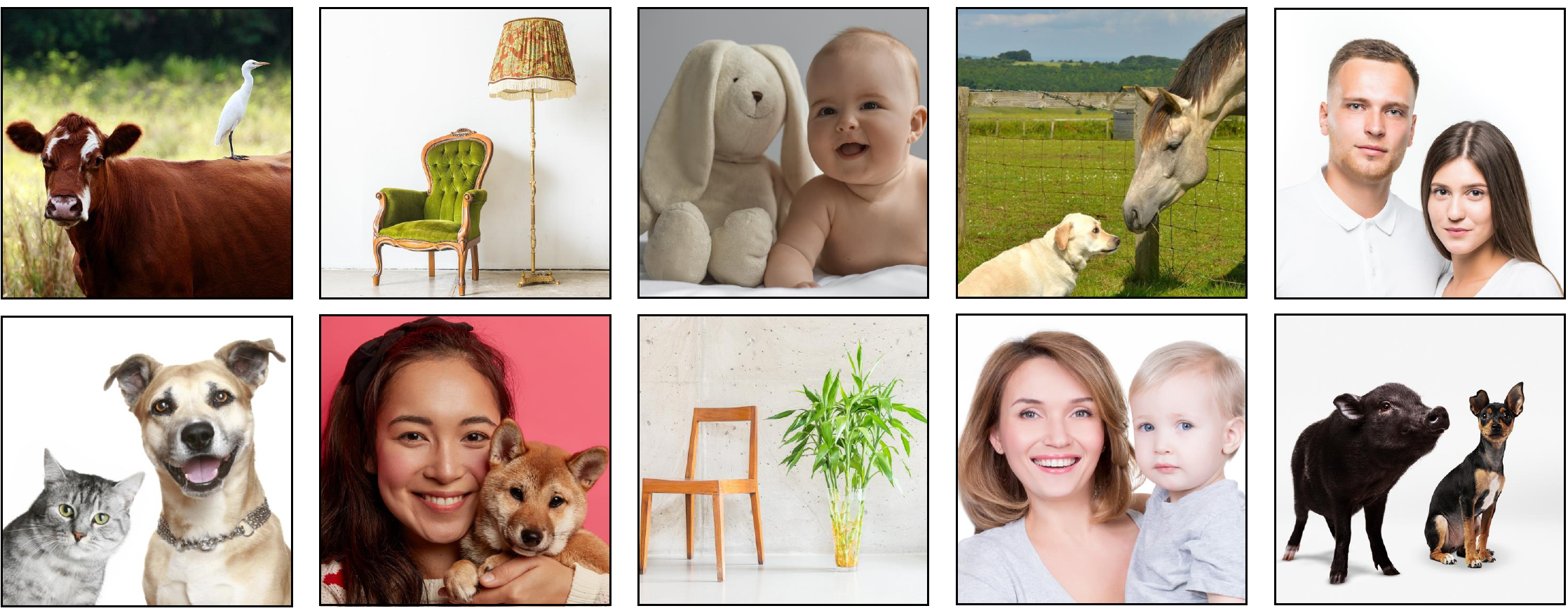}
    \caption{\textbf{Overview of ten datasets.} } 
    \label{figure:ten_datasets}
    \vspace{-15pt}
\end{figure*}

\myparagraph{Datasets.} We present each training image in \cref{figure:ten_datasets}.

\myparagraph{Textual Inversion \cite{gal2022image}.} We utilized the implementation from \cite{von-platen-etal-2022-diffusers} with 5000 training steps, a batch size of 4, and a learning rate of 0.0005. The input prompt, originally ``A photo of $V^*$'' in Textual Inversion, is modified to ``A photo of $V_1^*$ and $V_2^*$''.  The two new words ($V_1^*$ and $V_2^*$) are initialized with the classes from the input image. For example, if the image contains a cat and a dog, $V_1^*$ and $V_2^*$ token embeddings are initialized as the pre-trained ``cat'' and ``dog'' token embeddings.

\myparagraph{DreamBooth \cite{ruiz2023dreambooth}.} We employ the implementation from \cite{von-platen-etal-2022-diffusers} with 250 training steps, a batch size of 2, and a learning rate of $5\times 10^{-6}$. The input prompt is ``$V_1^*$ [class$_1$] and $V_2^*$ [class$_2$]'', consistent with our setting in \cref{section:preliminary}. Additionally, we generate 1000 ``a [class$_1$] and a [class$_2$]'' images using the pre-trained model \cite{ruiz2023dreambooth}. New modifiers are initialized as rare token embeddings.

\myparagraph{Custom Diffusion \cite{kumari2023multi}.} We employ the official implementation with 250 training steps, a batch size of 8, and a learning rate of $8\times 10^{-5}$.  The input prompt is also ``$V_1^*$ [class$_1$] and $V_2^*$ [class$_2$]'', and modifiers are also initialized as rare token embeddings. For regularization, 200 images are selected using clip-retrieval \cite{beaumont-2022-clip-retrieval} with the caption ``a [class$_1$] and a [class$_2$]''. We apply the default data augmentation in Custom Diffusion.

\myparagraph{DisenDiff (ours).} Implementation details are described in \cref{section:settings}. For the total loss in \cref{eq:total_loss}, the weight of $\mathcal{L}_{\text {bind}}$ is set to $0.01$ in all experiments. The weight of $\mathcal{L}_{\text {s\&s}}$ defaults to $0.01$ and occasionally adjusts to $0.001$ for specific cases.

\begin{table*}[!t]
    \centering
    \resizebox{\linewidth}{!}{
\begin{tabular}{llccccccccccc}
    \hline
    \multicolumn{1}{c}{}
& \textbf{Method}            & Cat+Dog                               & Cow+Bird                              & Man+Woman                    & Chair+Vase     & Chair+Lamp     & Dog+Pig        & Mother+Child   & Woman+Dog      & Horse+Dog      & Baby+Toy       & \textbf{Mean}  \\ \hline
& \textbf{Textual Inversion} & 0.732                                 & 0.656                                 & 0.550                        & 0.649          & 0.663          & 0.662          & 0.557          & 0.541          & 0.636          & 0.607          & 0.625          \\
& \textbf{DreamBooth}        & 0.732                                 & 0.701                                 & {\color[HTML]{000000} 0.606} & 0.815          & 0.784          & 0.701          & 0.601          & 0.625          & 0.708          & 0.689          & 0.696          \\
& \textbf{Custom Diffusion}  & 0.808                                 & 0.777                                 & 0.719                        & 0.811          & \textbf{0.798} & 0.771          & 0.705          & 0.706          & \textbf{0.747} & 0.775          & 0.762          \\
\cdashline{2-13}
    \multirow{-4}{*}{\begin{tabular}[c]{@{}l@{}}Image-alignment\\ (Mean)\end{tabular}}        & \textbf{Ours}              & {\color[HTML]{000000} \textbf{0.824}} & {\color[HTML]{000000} \textbf{0.783}} & \textbf{0.749}               & \textbf{0.822} & 0.795          & \textbf{0.773} & \textbf{0.718} & \textbf{0.737} & 0.744          & \textbf{0.808} & \textbf{0.775} \\ \hline
& \textbf{Textual Inversion} & 0.802                                 & 0.815                                 & \textbf{0.739}               & \textbf{0.814} & \textbf{0.834} & \textbf{0.834} & 0.764          & 0.776          & 0.811          & 0.767          & \textbf{0.796}          \\
& \textbf{DreamBooth}        & \textbf{0.804}                        & 0.816                                 & 0.738                        & 0.732          & 0.811          & 0.830          & \textbf{0.768} & \textbf{0.781} & 0.817          & 0.778          & 0.788 \\
& \textbf{Custom Diffusion}  & 0.773                                 & 0.843                                 & 0.731                        & 0.759          & 0.794          & 0.793          & 0.740          & 0.754          & 0.818          & \textbf{0.800} & 0.780          \\
\cdashline{2-13}
    \multirow{-4}{*}{\begin{tabular}[c]{@{}l@{}}Text-alignment\\ (Mean)\end{tabular}}         & \textbf{Ours}              & 0.774                                 & \textbf{0.847}                        & 0.727                        & 0.757          & 0.800          & 0.794          & 0.744          & 0.732          & \textbf{0.826} & 0.796          & 0.780          \\ \hline
& \textbf{Textual Inversion} & 0.743                                 & 0.690                                 & 0.527                        & 0.687          & 0.659          & 0.662          & 0.572          & 0.542          & 0.620          & 0.644          & 0.634          \\
& \textbf{DreamBooth}        & 0.736                                 & 0.774                                 & 0.679                        & 0.897          & 0.845          & 0.697          & 0.672          & 0.684          & 0.784          & 0.739          & 0.751          \\
& \textbf{Custom Diffusion}  & 0.856                                 & 0.843                                 & 0.801                        & \textbf{0.914} & \textbf{0.903} & 0.793          & 0.777          & 0.807          & \textbf{0.801} & 0.820          & 0.832          \\
\cdashline{2-13}
    \multirow{-4}{*}{\begin{tabular}[c]{@{}l@{}}Image-alignment\\ (Combined)\end{tabular}}    & \textbf{Ours}              & \textbf{0.865}                        & \textbf{0.855}                        & \textbf{0.828}               & 0.909          & 0.883          & \textbf{0.794} & \textbf{0.795} & \textbf{0.835} & 0.792          & \textbf{0.870} & \textbf{0.843} \\ \hline
& \textbf{Textual Inversion} & \textbf{0.797}                        & 0.805                                 & 0.738                        & \textbf{0.800} & \textbf{0.816} & \textbf{0.839} & 0.797          & 0.777          & 0.823          & 0.811          & \textbf{0.800} \\
& \textbf{DreamBooth}        & 0.780                                 & 0.823                                 & \textbf{0.762}               & 0.705          & 0.799          & 0.824          & \textbf{0.815} & \textbf{0.821} & \textbf{0.843} & \textbf{0.823} & 0.799          \\
& \textbf{Custom Diffusion}  & 0.736                                 & 0.882                                 & 0.719                        & 0.698          & 0.747          & 0.749          & 0.735          & 0.729          & 0.826          & 0.792          & 0.761          \\
\cdashline{2-13}
    \multirow{-4}{*}{\begin{tabular}[c]{@{}l@{}}Text-alignment\\ (Combined)\end{tabular}}     & \textbf{Ours}              & 0.747                                 & \textbf{0.896}                        & 0.708                        & 0.711          & 0.767          & 0.772          & 0.746          & 0.712          & 0.842 & 0.805          & 0.771          \\ \hline
& \textbf{Textual Inversion} & 0.756                                 & 0.688                                 & 0.527                        & 0.671          & 0.669          & 0.682          & 0.501          & 0.463          & 0.647          & 0.648          & 0.625          \\
& \textbf{DreamBooth}        & 0.763                                 & 0.697                                 & 0.545                        & 0.755          & 0.795          & 0.707          & 0.520          & 0.554          & 0.707          & 0.679          & 0.672          \\
& \textbf{Custom Diffusion}  & 0.818                                 & \textbf{0.768}                        & 0.661                        & 0.750          & \textbf{0.803} & \textbf{0.779} & \textbf{0.661} & 0.635          & \textbf{0.761} & 0.802          & 0.744          \\
\cdashline{2-13}
    \multirow{-4}{*}{\begin{tabular}[c]{@{}l@{}}Image-alignment\\ (Concept$_1$)\end{tabular}} & \textbf{Ours}              & \textbf{0.837}                        & 0.766                                 & \textbf{0.674}               & \textbf{0.766} & 0.804          & 0.771          & 0.651          & \textbf{0.678} & 0.752          & \textbf{0.808} & \textbf{0.751} \\ \hline
& \textbf{Textual Inversion} & 0.798                                 & 0.845                                 & 0.732                        & \textbf{0.830} & \textbf{0.840} & \textbf{0.822} & \textbf{0.729} & \textbf{0.738} & 0.815          & 0.771          & \textbf{0.792} \\
& \textbf{DreamBooth}        & \textbf{0.823}                        & 0.800                                 & 0.722                        & 0.768          & 0.792          & 0.818          & 0.724          & 0.715          & 0.819          & 0.775          & 0.779          \\
& \textbf{Custom Diffusion}  & 0.776                                 & \textbf{0.856}                        & \textbf{0.733}               & 0.812          & 0.805          & 0.777          & 0.688          & 0.718          & 0.819          & \textbf{0.831} & 0.781          \\
\cdashline{2-13}
    \multirow{-4}{*}{\begin{tabular}[c]{@{}l@{}}Text-alignment\\ (Concept$_1$)\end{tabular}}  & \textbf{Ours}              & 0.776                                 & \textbf{0.856}                        & 0.732                        & 0.809          & 0.809          & 0.774          & 0.693          & 0.677          & \textbf{0.822} & 0.826          & 0.777          \\ \hline
& \textbf{Textual Inversion} & 0.695                                 & 0.590                                 & 0.596                        & 0.589          & 0.660          & 0.642          & 0.598          & 0.619          & 0.640          & 0.528          & 0.616          \\
& \textbf{DreamBooth}        & 0.696                                 & 0.632                                 & 0.594                        & 0.795          & \textbf{0.711} & 0.699          & 0.612          & 0.635          & 0.635          & 0.650          & 0.666          \\
& \textbf{Custom Diffusion}  & 0.748                                 & 0.721                                 & 0.696                        & 0.769          & 0.688          & 0.741          & 0.675          & 0.675          & 0.678          & 0.702          & 0.709          \\
\cdashline{2-13}
    \multirow{-4}{*}{\begin{tabular}[c]{@{}l@{}}Image-alignment\\ (Concept$_2$)\end{tabular}} & \textbf{Ours}              & \textbf{0.770}                                 & \textbf{0.729}                        & \textbf{0.744}               & \textbf{0.790} & 0.699          & \textbf{0.754} & \textbf{0.708} & \textbf{0.697} & \textbf{0.688} & \textbf{0.747} & \textbf{0.733} \\ \hline
& \textbf{Textual Inversion} & \textbf{0.812}                        & \textbf{0.796}                        & \textbf{0.747}               & \textbf{0.817} & \textbf{0.847} & 0.842          & 0.766          & 0.812          & 0.794          & 0.719          & 0.795          \\
& \textbf{DreamBooth}        & 0.809                                 & 0.794                                 & 0.729                        & 0.725          & 0.843          & 0.848          & 0.765          & 0.808          & 0.790          & 0.737          & 0.785          \\
& \textbf{Custom Diffusion}  & 0.808                                 & 0.792                                 & 0.742                        & 0.767          & 0.830          & \textbf{0.853} & \textbf{0.797} & \textbf{0.815} & 0.808          & \textbf{0.775} & \textbf{0.799} \\
\cdashline{2-13}
    \multirow{-4}{*}{\begin{tabular}[c]{@{}l@{}}Text-alignment\\ (Concept$_2$)\end{tabular}}  & \textbf{Ours}              & 0.799                                 & 0.787                                 & 0.741                        & 0.752          & 0.823          & 0.836          & 0.792          & 0.807          & \textbf{0.814} & 0.757          & 0.791          \\ \hline
    \end{tabular}
    
}
    \vspace{-8pt}

    \caption{\textbf{Quantitative comparison on each dataset.} Evaluation metrics are outlined in Section \ref{section:settings} (higher is better for both metrics). We report four types of scores (Mean, Combined, Concept$_1$, Concept$_2$), and the averaged results across ten datasets are illustrated in Figure \ref{figure:main_comparsion}. The term ``Cat+Dog'' signifies the presence of both ``Cat'' and ``Dog'' concepts within the dataset.}
    \label{table:quantitative_results_each_dataset}
    \vspace{-5pt}
\end{table*}

\end{document}